\documentclass[runningheads]{llncs}
\usepackage{graphicx}
\usepackage[utf8]{inputenc}
\usepackage{amssymb}
\usepackage{multirow}
\usepackage{wrapfig}
\begin{document}
%
\title{Learning to Describe Player Form in the MLB}
%
%
\author{Connor Heaton \and Prasenjit Mitra}
\authorrunning{Heaton et al.}
%

\institute{The Pennsylvania State University, State College, PA 16802, USA \\
\email{\{czh5372,pmitra\}@psu.edu}}

\maketitle              
\begin{abstract}

Major League Baseball (MLB) has a storied history of using statistics to better understand and discuss the game of baseball, with an entire discipline of statistics dedicated to the craft, known as \textit{sabermetrics}. At their core, all \textit{sabermetrics} seek to quantify some aspect of the game, often a \textit{specific} aspect of a player's skill set - such as a batter's ability to drive in runs (RBI) or a pitcher's ability to keep batters from reaching base (WHIP). 
While useful, such statistics are fundamentally limited by the fact that they are derived from an account of \textit{what} happened on the field, not \textit{how} it happened.
As a first step towards alleviating this shortcoming, we present a novel, contrastive learning-based framework for describing player \textit{form} in the MLB. We use \textit{form} to refer to the way in which a player has impacted the course of play in their recent appearances. Concretely, a player's \textit{form} is described by a 72-dimensional vector. By comparing clusters of players resulting from our \textit{form representations} and those resulting from traditional \textit{sabermetrics}, we demonstrate that our \textit{form representations} contain information about how players impact the course of play, not present in traditional, publicly available statistics. We believe these embeddings could be utilized to predict both in-game and game-level events, such as the result of an at-bat or the winner of a game.



\keywords{Machine Learning \and Player Valuation}
\end{abstract}
\section{Introduction}

As the sport of baseball grew in popularity, fans, players, and managers desired a more pointed way of discussing, and arguing over, the game. The more mathematically inclined fans realized that they could use statistics to describe how players have historically performed, and how those performances have translated in to advantages for their team - and thus, \textit{sabermetrics} was born\footnote{A somewhat simplified history}.

While \textit{sabermetrics} have undoubtedly changed how players, fans, and front offices alike interact with the game - introducing new statistics, such as WAR, OPS+, SIERRA, and BABIP among others, and inspiring new strategies, such as the defensive infield shift, offensive launch angles, etc. - such statistics are fundamentally limited in their descriptive power by the fact that they are derived from an account of \textit{what} happened on the field, not \textit{how} it happened. 


To see why a description of \textit{what} happened is less desirable than a description of \textit{how} it happened, consider two at-bats: one between \textit{pitcher A} \& \textit{batter B}, and another between \textit{pitcher X} \& \textit{batter Y}. In the former, \textit{pitcher A} got ahead in the count 0-2, but \textit{batter B} battled back to a full-count, eventually hitting a ball to deep right-field and reaching first base comfortably. For the latter, \textit{pitcher X} fell behind 2-0 in the count, before \textit{batter Y} hit a dribbler down the third base line and beat the throw to first. The most simple way of describing the two at bats would be to say a single was recorded in both cases, which would do nothing to differentiate between the two at bats. Information could be included about how many pitches were thrown in each at-bat, but that still doesn't tell the whole story. For example, it wouldn't convey that \textit{batter B} was able to battle back from an 0-2 count and reach base or the way in which either pitcher sequenced their pitches. Furthermore, it would not convey that \textit{batter B} had power-hit a fly-ball to deep-right nor that \textit{batter Y} had the speed to beat out the throw to first.



Typically, \textit{sabermetrics} have been used to describe some aspect of a player's game over a relatively large time-scale. Intuitively, in a statistical sense, this makes sense - they are statistics derived from a sample population, and the larger the sample population, the more \textit{accurate} the computed sample statistic will be with respect to describing the population at large. The interpretations of these \textit{sabermetrics} often ``break down'' when working with a small number of samples. For example, it would not make much sense to use batting average or WHIP to describe the players participating in the at-bats mentioned above. 
\textit{Pitcher A} and \textit{pitcher B} will have a WHIP of $\infty$ while \textit{batter X} and \textit{batter Y} will have a batting average of 1.000. They aren't incorrect, however - they accurately describe \textit{what} happened, but do little to reveal \textit{how} it happened. For this reason, we believe it is sub-optimal to derive a description of a player's short-term performance using traditional \textit{sabermetrics}.



In 2015, Statcast systems were added to all 30 MLB stadiums~\cite{casella2015statcast}. 
These systems record highly detailed information about many aspects of the game including player positioning in the field, thrown pitch types, pitch velocity, pitch rotation, hit distance, batted ball exit velocity, and launch angle, among others, for every pitch thrown in every game. Analysis of this Statcast data has already influenced how the game is understood, drawing more attention to batters' launch angle, for example. We believe new insights can be found by analyzing Statcast data as a sequence of records instead of records in isolation. 


\section{Related Work}

For an extended description of the many \textit{sabermetrics}, we direct the interested reader toward \textit{Understanding sabermetrics} by Costa, Huber, and Saccoman. Below, we describe similar work towards obtaining player representations and related work in machine learning (ML).

\subsection{(batter$\vert$pitcher)2vec}
The (batter$\vert$pitcher)2vec model was proposed in 2018, motivated by recent advances in natural language processing (NLP)~\cite{alcorn20162vec}. Player embeddings were learned by modeling at-bats - Given an ID for the batter and pitcher taking part in an at-bat, the model was asked to learn embeddings that describe each player and can be used to predict the result of said at-bat. The model was trained using MLB at-bats from 2013 through 2016. Once an embedding was learned for each player, the author demonstrated how they can be used to make predictions as to the result of an unseen at-bat with more accuracy than previous methods.

\subsection{Transformers, BERT, \& Image-GPT}
The \textit{transformer} architecture rose in popularity thanks in large part to its use in the BERT language model~\cite{vaswani2017attention}\cite{devlin2018bert}. The motivating principal behind BERT is the notion that the meaning of words can be inferred by analyzing the context in which they naturally appear, and the transformer architecture, along with a special training regimen, enable BERT to do just that. 

To learn the language, BERT browses the internet and performs two tasks when it comes across a piece of text: 1) Masked Language Modeling (MLM) and 2) Next Sentence Prediction (NSP). MLM is essentially BERT creating fill-in-the-blank questions for itself. For example, if BERT comes across the text ``I love you,'' it may create a fill-in-the-blank question in the form of ``I love \_\_\_\_.'' By analyzing the context surrounding the blank, the model is likely to fill the blank with ``you.'' Instead, if the fill-in-the-blank question were ``I go to the gym every day. I love \_\_\_\_,'' the context may induce the model to fill the blank with ``exercise.'' By learning to fill in the blank correctly, BERT is learning to infer from the context the meaning associated with different words.

The NSP task helps BERT learn the emergent meaning associated with various sequences 
of characters. Given two sentences, the model is asked to make a binary prediction as to whether or not these sentences appeared next to each other ``in the wild.'' For example, if the example above was separated and given to the model as two sentences in ``I go to the gym every day'' and ``I love exercise,'' the model would be expected to respond affirmatively. By repeatedly performing this test, the model will begin to understand the semantics emerging from the sequence of words and characters - if you do something every day, you likely ``love'' it, and the ``gym'' is associated with ``exercise,'' for example.

Upon seeing the success BERT and similar models had working with natural language, Chen et al. noted that their MLM training objective closely resembled that of Denoising Autoencoders, which were originally designed
to work with images, and explored the extent to which transformers could be used to learn image representations using a similar training scheme~\cite{chen2020generative}. Instead of learning to ``fill in the blank'' as BERT would, this model, dubbed \textit{Image-GPT}, would learn to impute the missing pixels in a corrupted image. In much the same fashion that BERT understands language as a collection of characters and words and learns their meaning by analyzing the context in which they appear, \textit{Image-GPT} perceives images as a collection of pixels with varying intensities of red, green, and blue, and learns the role they play in the emergent semantics of the image by analyzing the context in which they appear.



BERT and \textit{Image-GPT} demonstrate that when paired with an appropriate training objective, transformers can be effective learners of atomic-element representations by leveraging the context in which these atomic-elements appear. Here, we use \textit{atomic-element} to refer to the lowest unit of information the model is capable of expressing or understanding - for BERT, groups of English characters are the \textit{atomic-elements}, while pixel values are the \textit{atomic-elements} for \textit{Image-GPT}. Furthermore, they demonstrate how the same model that learned these atomic-element representations also learns how to discern an emergent meaning when these representations are viewed in conjunction with one another.

\subsection{Contrastive Learning}
Contrastive Learning is a training scheme often used in Computer Vision (CV) applications where a model's training objective is to minimize a contrastive loss objective among a batch of sample records~\cite{chen2020simple}.  The motivating theory behind contrastive learning is that similar inputs should result in similar outputs from a representation-learning model - in our application, similar sequences of at-bats should be described with similar \textit{form} vectors. When learning via self-supervised contrastive loss, for example, the model is given two different \textit{views} of a single image, and encouraged to produce similar outputs~\cite{misra2020self}. Furthermore, the output produced by two \textit{views} of the same image are expected to be dissimilar to outputs resulting from views of different source images. The different \textit{views} are often obtained by a combination of randomly cropping, rotating, resizing, inverting, or otherwise distorting the source image. 

The self-supervised contrastive loss objective is given in equation \ref{con_loss_eq}, where $I \equiv \{0, 1, ..., 2N-1\}$ is the set of indices in the batch, $j(i)$ is the \textit{positive} sample(s) for record \textit{i}, $\cdot$ is the inner product, $ \tau \in \mathbb{R}^+$ is a scaling temperature value, and $a \in A(i)$ is the set including the \textit{positive} and \textit{negative} samples for record $i$ \cite{khosla2020supervised}. Record \textit{j} is a considered \textit{positive} sample for record \textit{i} if it is derived from the same source image, otherwise it is considered a \textit{negative} sample.

\begin{equation}\label{con_loss_eq}
    L^{self} = \sum_{i \in I}L_i^{self} = -\sum_{i \in I}log\frac{exp(z_i \cdot z_{j(i)} / \tau)}{\sum_{a \in A(i)}exp(z_i \cdot z_a / \tau)}
\end{equation}

\section{Our Method}
The principal motivation behind our work towards describing player \textit{form} is very much the same as that of contrastive learning - players who impact the game in similar ways should be described using similar \textit{form} vectors.
We do not have ground truth player \textit{form} labels (vectors), but we do know the same batter at two very close points in time should be described similarly. 
In the sections that follow, we describe how data was collected, our player \textit{form} model was trained, and discrete player forms were obtained. 

\subsection{Data Collection and Organization}
While data was originally collected by the Statcast system, we use the Python package \texttt{pybaseball}\footnote{https://github.com/jldbc/pybaseball} to collect data used for our study and populate a local \texttt{sqlite3} database. We collected two types of data using this package: 1) pitch-by-pitch data and 2) season-by-season statistics. 
Pitch-by-pitch data was collected for the 2015 through 2018 seasons, and contains information such as pitch type, batted ball exit velocity, and launch angle among others.
Season-by-season statistics were collected for the 1995 season through 2018, and contain position-agnostic information such as WAR and age in addition to position-specific information such as WHIP for pitchers and OPS for batters. 
\begin{wraptable}{r}{4cm}
    \begin{center}
    \begin{tabular}{| p{2cm} | p{1cm} |}
        \hline
        \# Games & 9,860 \\
        \hline
        \# PA & 750k \\
        \hline
        \# Pitches & 2.9M \\
        \hline
        \# Batters & 1,690 \\
        \hline
        \# Pitchers & 1,333 \\
        \hline
    \end{tabular}
    
    \label{dataset_summary}
    \caption{Dataset summary}
    \end{center}
\end{wraptable}

Each record in our pitch-by-pitch table is accompanied by three key values- 1) \textit{game\_pk}, 2) \textit{AB\_number}, and 3) \textit{pitch\_number}. The \textit{game\_pk} is a unique value associated with each game played in the MLB. Within each game, each at-bat has a corresponding \textit{AB\_number}, and each pitch thrown in an at-bat an associated \textit{pitch\_number}. By using these three pieces of information, we can completely reconstruct the sequence of events which constitute an MLB game. A summary of our collected dataset is given in table \ref{dataset_summary}.


\subsection{Describing In-game Events}\label{desribing_events}
Typically, one would use terms like \textit{single}, \textit{home run}, or \textit{strikeout} to describe the outcome of an at bat. Using this terminology to describe the outcome of at-bats to our model would be insufficient, however, as it tells an incomplete story. For example, did other runners advance on the play?

For this reason, we describe the outcome of a pitch in terms of the \textbf{change} in the \textit{gamestate}, where the \textit{gamestate} refers to 1) ball-strike count, 2) base occupancy, 3) number of outs, and 4) score. These changes in \textit{gamestate} will constitute the vocabulary that our model will learn to understand. In total, we identify 325 possible changes in \textit{gamestate} and the result of any thrown pitch can be described by one of these changes in \textit{gamestate}. We colloquially refer to these \textit{gamestate} changes as \textit{gamestate deltas}. From our pitch-by-pitch table, we also have information describing the thrown pitch
and batted ball which induced this change in the game state, such as pitch type, pitch rotation, location over the plate, and batted ball distance and launch angle among others.

In aggregate, the \textit{gamestate deltas} describe \textit{what} happened and \textit{how}, but do not describe \textit{who} was involved. 
We describe the pitcher, batter, and historical matchup between the two using traditional sabermetrics.

\begin{wraptable}{l}{5.28cm}
    \begin{center}
    \begin{tabular}{| c | c | c | c|}
        \hline
          & Batter & Pitcher & Matchup \\ 
        \hline
        Career & 167 & 141 & 137 \\
        \hline
        Season & 137 & 137 & 137 \\
        \hline
        Last 15 & 137 & 137 & N/A \\
        \hline
        This Game & 137 & 137 & 137 \\
        \hline
    \end{tabular}
    
    \label{supplemental_data}
    \caption{Supplemental statistics at different time scales.}
    \end{center}
\end{wraptable}

We use statistics derived from four different temporal scales when describing the pitcher, batter, and matchup between the two. A summary of these supplemental features are given in table \ref{supplemental_data}

When presenting this information to the model, the 1,541 supplemental features are projected to a lower dimension such that roughly half the data at each input index describes the \textit{gamestate delta} and the other half 
describes the \textit{players} involved in the at-bat, thrown/batted ball, and stadium.

\subsection{Player Form Learning}
We seek to describe player \textit{form} - how a player has impacted the game in their recent appearances - so we must identify a \textit{window} of activity (consecutive appearances) we wish to describe for each player. 
Once this \textit{window} is identified, we can then create two \textit{views} (sets of consecutive appearances) of the player's influence on the game in this \textit{window} of activity. These two \textit{views} describe the same player over a relatively small period of time; so they should induce similar outputs from our player \textit{form} model. 
Furthermore, \textit{views} from the same \textit{window} of activity for the same player should be dissimilar to \textit{views} derived from \textit{windows} of activity of other players, and even other \textit{windows} for the same player.

For batters, we define a \textit{window} of activity as a sequence of 20 consecutive at-bats for that batter. Then, for each \textit{window} of activity, we derive the first \textit{view} as the first 15 at-bats in the \textit{window}, and the second \textit{view} as the final 15 at-bats in the window. For pitchers, we define a \textit{window} of activity as a sequence of 100 consecutive at-bats for which they pitched, and a \textit{view} as 90 at-bats. That is, the first 90 at-bats in the \textit{window} serve as the first \textit{view} while the final 90 at-bats serve as the second \textit{view}. Batters have an average 4.2 plate appearances per game\footnote{https://fivethirtyeight.com/features/relievers-have-broken-baseball-we-have-a-plan-to-fix-it/}, and starting pitchers face an average of 23.3 batters per start from 2015-2018\footnote{https://blogs.fangraphs.com/starting-pitcher-workloads-have-been-significantly-reduced-in-2020/}, so \textit{view} sizes were selected such that each view covered \textit{roughly} four games per player.

Present in each input sequence will also be a special \textit{[CLS]} token, which will be used in a similar fashion as BERT's \textit{[CLS]} token. That is, our model will learn to process the input data such that the processed \textit{[CLS]} embedding will sufficiently describe the entirety of the input.

Our model describes players over a short period of time, i.e.,
15 at-bats, while (batter$\vert$pitcher)2vec describes players over a much larger time scale, four seasons. However, we would still be able to describe players over a much larger time-scale by viewing consecutive sequences of 15 at-bats in the aggregate, making our model much more versatile - the same model can be used to derive a description of a player over the course of 15 at bats, or four seasons.

\subsubsection{Model Architecture} 
We use a multi-layer, bidirectional transformer encoder, based on the original implementation used in BERT~\cite{vaswani2017attention}\cite{devlin2018bert}. Our model consists of 8 transformer layers, 8 attention heads in each layer, and a model dimension of 512. Our model learns embeddings to describe many aspects of the input data, including \textit{gamestate deltas}
, stadiums
, player positions
, pitch types
, and pitch locations over the plate
. The remaining information at each input index is derived from a two-layer projection of the supplemental player inputs, described in section \ref{desribing_events}, and real-valued attributes of the thrown pitch and batted ball.

Additionally, the model learns embeddings, which help position the inputs with respect to one another, such as the at-bat number within the \textit{window} and the pitch number within said at-bat. Separate models are trained to describe pitcher and batter \textit{forms} with no shared weights.


\subsubsection{Training} We use two tasks to train our model: 1) Masked Gamestate Modeling (MGM) and 2) Self-supervised Contrastive Learning. The MGM task is akin to MLM, with roughly 15\% of \textit{gamestate delta} tokens in the input sequence masked and the model asked to impute the missing values. In addition to learning the relation between \textit{gamestate delta} tokens - e.g., three consecutive balls are often shortly followed by a fourth - the model also learns the relation between different types of batters and pitchers participating in the at-bat. For example, if the supplemental inputs describe a shutdown pitcher, poor batter, and pitcher-friendly stadium, the corresponding \textit{gamestate delta} is likely to be to the pitcher's benefit.

\begin{figure*}[ht!]
    \centering
    \includegraphics[scale=0.275]{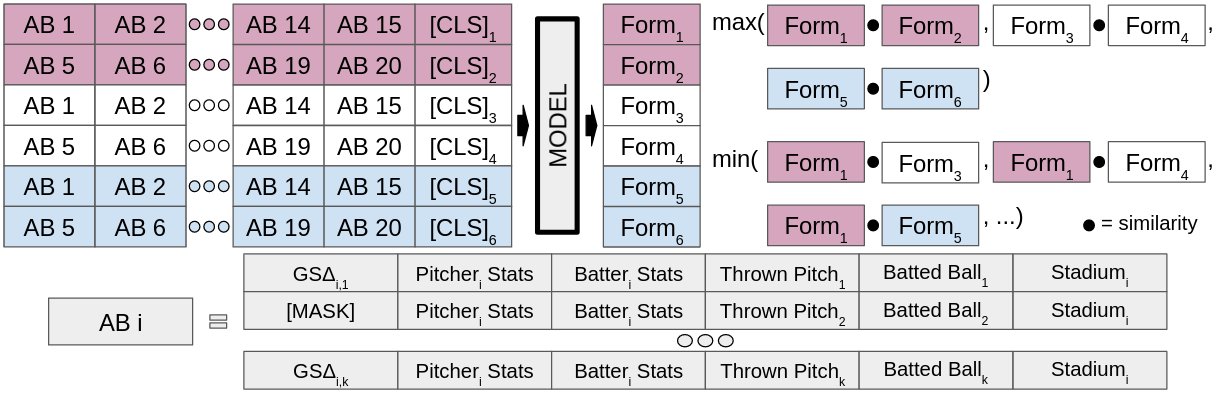}
    \caption{Example of how our model processes a batch of data while learning to describe batters. Each record in the batch consists of 15 at-bats and a special \textit{[CLS]} token, and each at-bat consists of one or more pitches. For each pitch, the model is given statistics describing the pitcher and batter involved, metrics describing thrown pitch type and batted ball (when applicable), and embeddings describing the stadium and the resulting \textit{gamestate delta}. The model is asked to predict the masked \textit{gamestate delta} tokens using the context in which the masked token appears. Once processed by the model, the embedding for the \textit{[CLS]} assumed to describe the 15 corresponding at-bats. These embeddings are projected to a 72-dimension space before being used to compute the self-supervised contrastive loss.}
    \label{overview}
\end{figure*}

The self-supervised contrastive learning task is used to train our model to induce representations that are similar for \textit{views} from the same \textit{window} and dissimilar for \textit{views} of from different \textit{windows}. Concretely, our model produces a 72-dimensional representation for each \textit{view}, which is used in computing the self-supervised contrastive loss. An example of how the model processes a batch of inputs is presented in figure \ref{overview}.


We use an Adam optimizer with $\beta_1=0.9$, $\beta_2=0.999$, and learning rate of $5e^{-4}$. When learning to describe batters, our model is trained using a batch size of 78 for 90,000 iterations, with 7,500 of the iterations being warm-up. A pitchers' \textit{form} is described by a larger number of at-bats, so we train our pitcher model using a batch size of 36 for 35,000 iterations with 2,500 iterations of warm-up. 

\subsection{Discretizing Player Forms}
We compute a \textit{form} representation for all players in the starting lineup at game-start for all regular season games from 2015-2018. 
Then, to identify players who have impacted the game in a similar capacity at various points in time, we perform agglomerative clustering with Ward linkage on the form representations to obtain discrete form ID's~\cite{rokach2005clustering}. 
For a point of comparison, we follow a similar clustering process using traditional \textit{sabermetrics} to describe players. 
That is, the players' corresponding supplemental inputs, mentioned in \ref{desribing_events}, without the \textit{in-game} split. We perform Principal Component Analysis on the statistics used to describe each type of player  prior to clustering.


\section{Results}
While a more thorough analysis is required to better understand the representations produced by our model, comparing clusters of players derived from traditional sabermetrics and from \textit{form representations} can give an intuition as to the information contained in the representations. 

\begin{figure*}[ht!]
    \centering
    \includegraphics[scale=0.4]{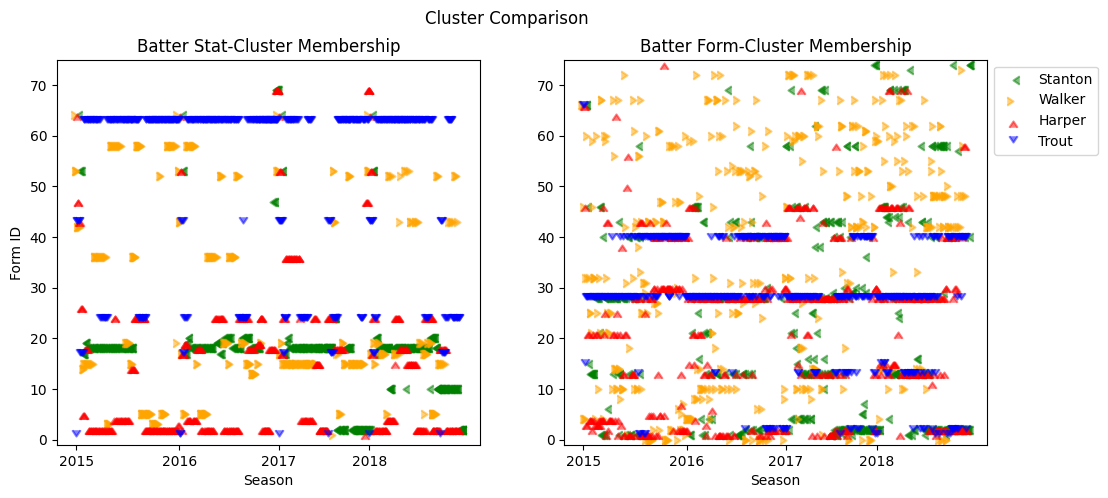}
    \caption{Discrete batter forms at game-start for games in the 2015 through 2018. 
    }
    \label{batter_cluster_comp}
\end{figure*}

Figure \ref{batter_cluster_comp} presents a comparison of cluster membership over time for four batters: Bryce Harper, Mike Trout, Giancarlo Stanton, and Neil Walker. Harper and Trout are somewhat similar, high impact outfielders, while Stanton can tend to be more of a streaky power hitter, and Walker is perhaps more of a utility infielder. In analyzing the plot describing \textit{stat-clusters} in figure \ref{batter_cluster_comp}, we see minimal overlap between Harper and Trout. This is an undesirable representation of form, as in our estimation, Harper and Trout tend to impact the game in a similar way.
Conversely, we seem to notice a strong association between Harper and Trout in the plot describing their \textit{form-clusters} in figure \ref{batter_cluster_comp}.

\begin{figure*}[ht!]
    \centering
    \includegraphics[scale=0.4]{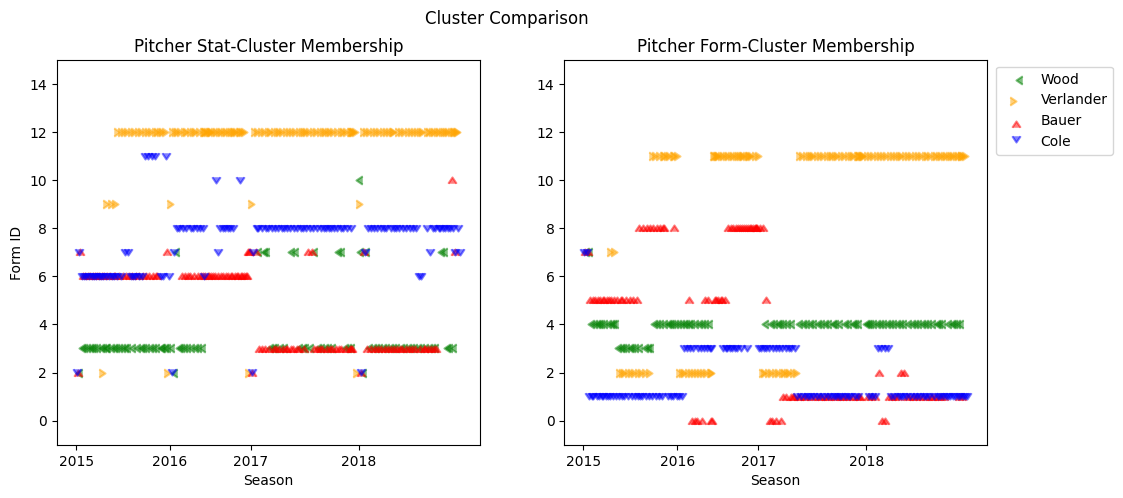}
    \caption{Discrete pitcher forms at game-start for games in the 2015 through 2018. 
    }
    \label{pitcher_cluster_comp}
\end{figure*}

Figure \ref{pitcher_cluster_comp} presents a comparison of cluster membership over time for four starting pitchers: Gerrit Cole, Alex Wood, Trevor Bauer, and Justin Verlander. Cole had rather poor performances throughout 2016 and some of 2017 before snapping back in to All-Star form with Houston in 2018 and 2019. In looking at Cole's \textit{stat-clusters} in figure \ref{pitcher_cluster_comp}, we see that he is consistently mapped to cluster 8 from 2016 onward. This would seem to be an undesirable to describe how he impacted the game, then, as he clearly impacted the game in very different ways in 2016 \& 2017 versus 2018. While at the moment we cannot say for certain what signal is contained in our \textit{form representations} we see they describe 2016 Cole differently than 2018 Cole. 




\section{Conclusion \& Future Work}
We believe this work serves as a strong starting point in a line of work towards a new way of describing how MLB players - and athletes in other sports - impact the course of play, in a manner not contained in existing, publicly available statistics. Moving forward, we would like to gain a stronger understanding of the contents of the produced \textit{form representations} and how they can be leveraged towards specific ends, such as predicting the result of an at-bat, the winner of an MLB game, or the occurrence of an injury. Furthermore, it would be interesting to take a closer look at the embeddings learned by our model. It would be interesting to explore the relation between different stadiums, for example, from the perspective of both the batter and pitcher.

%
%
\bibliographystyle{splncs04}
\bibliography{mybibliography}

\begin{thebibliography}{1}
\providecommand{\url}[1]{\texttt{#1}}
\providecommand{\urlprefix}{URL }
\providecommand{\doi}[1]{https://doi.org/#1}

\bibitem{alcorn20162vec}
Alcorn, M.A.: 2vec: statistic-free talent modeling with neural player
  embeddings. MIT Sloan Sports Analytics Conference (2016)

\bibitem{casella2015statcast}
Casella, P.: Statcast primer: Baseball will never be the same.
  \url{https://www.mlb.com/news/statcast-primer-baseball-will-never-be-the-same/c-119234412}
  (2015), [Online; accessed 20-June-2021]

\bibitem{chen2020generative}
Chen, M., Radford, A., Child, R., Wu, J., Jun, H., Luan, D., Sutskever, I.:
  Generative pretraining from pixels. In: International Conference on Machine
  Learning. pp. 1691--1703. PMLR (2020)

\bibitem{chen2020simple}
Chen, T., Kornblith, S., Norouzi, M., Hinton, G.: A simple framework for
  contrastive learning of visual representations. In: International conference
  on machine learning. pp. 1597--1607. PMLR (2020)

\bibitem{devlin2018bert}
Devlin, J., Chang, M.W., Lee, K., Toutanova, K.: Bert: Pre-training of deep
  bidirectional transformers for language understanding. arXiv preprint
  arXiv:1810.04805  (2018)

\bibitem{khosla2020supervised}
Khosla, P., Teterwak, P., Wang, C., Sarna, A., Tian, Y., Isola, P., Maschinot,
  A., Liu, C., Krishnan, D.: Supervised contrastive learning. arXiv preprint
  arXiv:2004.11362  (2020)

\bibitem{misra2020self}
Misra, I., Maaten, L.v.d.: Self-supervised learning of pretext-invariant
  representations. In: Proceedings of the IEEE/CVF Conference on Computer
  Vision and Pattern Recognition. pp. 6707--6717 (2020)

\bibitem{rokach2005clustering}
Rokach, L., Maimon, O.: Clustering methods. In: Data mining and knowledge
  discovery handbook, pp. 321--352. Springer (2005)

\bibitem{vaswani2017attention}
Vaswani, A., Shazeer, N., Parmar, N., Uszkoreit, J., Jones, L., Gomez, A.N.,
  Kaiser, L., Polosukhin, I.: Attention is all you need. arXiv preprint
  arXiv:1706.03762  (2017)

\end{thebibliography}

\end{document}